\documentclass[conference, 12pt, onecolumn]{IEEEtran}
\usepackage{blindtext}
\usepackage{titlesec}
\title{Sections and Chapters}
\usepackage{cite}
\usepackage{amsmath,amssymb,amsfonts}
\usepackage{algorithmic}
\usepackage{graphicx}
\usepackage{textcomp}
\usepackage{xcolor}
\usepackage{amsfonts}       
\usepackage{nicefrac}       
\usepackage{microtype}      
\usepackage{wrapfig,lipsum,booktabs}
\usepackage{soul}
\usepackage{multirow}
\usepackage{lscape}
\usepackage{subfigure}
\usepackage{adjustbox}
\usepackage{enumitem}
\usepackage{tabularx}
\usepackage{caption}
\usepackage{fancyhdr}
\usepackage{url}
\usepackage{booktabs}

\usepackage{mathtools}
\usepackage[ruled,vlined]{algorithm2e}
\usepackage{float}
\usepackage{siunitx}
\usepackage{circledsteps}
\usepackage{setspace}
\usepackage[margin=1in]{geometry}
\usepackage{colortbl}
\definecolor{Gray}{gray}{0.85}
\definecolor{aliceblue}{rgb}{0.94, 0.97, 1.0}
	\definecolor{beaublue}{rgb}{0.74, 0.83, 0.9}
\definecolor{blond}{rgb}{0.98, 0.94, 0.75}
\definecolor{beige}{rgb}{0.96, 0.96, 0.86}
	\definecolor{cornsilk}{rgb}{1.0, 0.97, 0.86}
	\definecolor{platinum}{rgb}{0.9, 0.89, 0.89}
\definecolor{lavendermist}{rgb}{0.9, 0.9, 0.98}
\definecolor{oldlace}{rgb}{0.99, 0.96, 0.9}
\def\BibTeX{{\rm B\kern-.05em{\sc i\kern-.025em b}\kern-.08em
    T\kern-.1667em\lower.7ex\hbox{E}\kern-.125emX}}
\begin{document}

\title{RealDiffFusionNet: Neural Controlled Differential Equation Informed Multi-Head Attention Fusion Networks for Disease Progression Modeling Using Real World Data
}

 \author{\IEEEauthorblockN{Aashish Cheruvu
}
\IEEEauthorblockA{
Central Bucks High School South\\
Warrington, PA, USA\\
aashcheruvu@gmail.com
}
\and
\IEEEauthorblockN{Nathaniel Rigoni}
\IEEEauthorblockA{
Lockheed Martin\\
Huntsville, AL, USA \\
nathaniel.c.rigoni@us.lmco.com}
}


\maketitle
\newpage

\begin{abstract}

The understanding of disease progression is at the heart of effective healthcare, offering invaluable insights into the course of illnesses, enabling timely interventions, and ultimately enhancing the quality of patient care. Electronic health records have brought us closer to harnessing real-world data to gain these insights. However, the volume and complexity of real-world data warrants the use of powerful techniques like deep learning. This paper presents a novel deep learning-based approach named RealDiffFusionNet that incorporates Neural Controlled Differential Equations (Neural CDE) - a class of time series models that are robust in handling irregularly sampled data - and multi-head attention as a mechanism to incorporate and align relevant multimodal context (image data, time invariant data, etc.) at each time point. Long short-term memory (LSTM) models were also used as a baseline to compare against Neural CDE performance. Two different datasets were used in the current study. One of the datasets was from the Open-Source Imaging Consortium (OSIC) containing structured time series data of the lung function with a baseline CT scan of the lungs. The second dataset used was from Alzheimer's Disease Neuroimaging Initiative (ADNI), that included a series of MRI scans along with demographics, physical examinations, and cognitive assessment data. An ablation study was performed to understand the role of CDEs, multimodal data, attention fusion, and interpolation strategies on model performance. When the baseline models (without attention) were evaluated, the use of multimodal data resulted in an improvement in Neural CDE performance, with a lower test RMSE (0.5405). Additionally, the performance of multimodal Neural CDE was also superior compared to multimodal LSTM (1.396). In the attention-based architectures, fusion through concatenation and rectilinear interpolation were found to improve model performance. The performance of the proposed RealDiffFusionNet was found to be superior (0.2570) to all models. For the ADNI dataset, between the Neural-CDE and LSTM models trained only on the structured data, the test RMSE were comparable (0.471 for LSTM vs. 0.4581 Neural-CDE). Furthermore, the addition of derived image features from patients’ MRI series to the structured data (multimodal) resulted in an improvement in Neural-CDE performance, with a lower test RMSE (0.4372 with multimodal data Vs 0.4581 with structure data). RealDiffFusionNet has shown promise in utilizing CDEs and multimodal data to accurately predict disease progression.

\end{abstract}

 \newpage
\tableofcontents

\pagestyle{fancy}
 \newpage
\section{Introduction}

Optimizing disease management relies heavily on the capability to forecast disease progression both promptly and precisely. 
For example, in the case of cancer, identifying the early stages of the disease and disease progression through various stages (e.g., Stage I-IV), as well as predicting patient outcomes at an early stage are helpful in choosing the appropriate interventions\cite{greene2008staging}.
To understand the disease and its progression, a comprehensive and thorough utilization of data from various sources (e.g., Electronic Health Records (EHRs), etc.) is the most important first step\cite{ma2021advances,miotto2018deep}. 
However, given the rapidly expanding array of tools and the surge in data volumes, traditional heuristics-based prognoses have become unfeasible for physicians\cite{mcginnis2013imperative}. To this end, Machine Learning (ML) methods that can process huge amounts of data in multiple dimensions are the most promising tools for assisting physicians. ML algorithms can discover, classify, and identify patterns and relationships between various disease characteristics and effectively predict future outcomes of disease\cite{JAVAID202258}.

 Long Short-Term Memory neural networks, or LSTMs, are a members of Recurrent Neural Networks that can model sequential characteristics by explicitly allowing the model to selectively choose useful information to pass to the next hidden state\cite{hochreiter1997long}. 
This is done with multiple gates that selectively add and remove information from the cell state. 
Various gates, cell states approaches are used to address the shortcomings of the RNNs. 
However, for RNNs including LSTM there is a need for the data to be discrete and demand that measurements be taken at steps of constant time intervals – known as regular sampling\cite{lipton2015critical}. 
However, most commonly, data in the medical domain is multivariate (multiple features that change with time), is irregularly sampled (non-uniform time intervals between observations), and unaligned (data for different features are collected at different times)\cite{zhang2022graphguided}.

A class of models known as Neural Differential Equations (NDEs) can address these issues related to irregular sampling and data alignment across different covariates\cite{chen2018neural}. 
NDEs use a neural network to model change in the data through a differential equation, which allows them to model sequences in a continuous manner. 
Because these functions model data continuously with respect to time, measurements at regular time intervals are not needed. 
Neural-controlled differential equations (Neural CDEs) are an extension of NDEs specifically designed to handle irregularly sampled multivariate time-series data\cite{kidger2020neural}. 
In addition, Neural CDEs offer multiple advantages including the ability to process incoming data, (which may even be partially observed) and can also be trained with memory-efficient adjoint-based backpropagation across observations.

The introduction of attention mechanisms has sparked a transformative shift, with the medical field experiencing particularly groundbreaking applications\cite{NIU202148}. By equipping computational models with the ability to selectively focus on pertinent aspects of complex data, attention mechanisms mirror a cognitive process akin to human focus, enhancing the model's performance in tasks that require precision and nuanced understanding. From analyzing medical images to parsing patient records and literature, attention-based models are redefining the possibilities in medical research and clinical practice, promising significant strides towards personalized and predictive medicine \cite{NIPS2017_3f5ee243}. 

Real-world data has emerged as a pivotal asset in revolutionizing healthcare practices, offering insights that were previously inaccessible\cite{katkade2018real}. This data encompasses information derived from diverse sources such as electronic health records, patient-generated data, wearable devices, and claims databases. Its application in healthcare is far-reaching, from enhancing clinical decision-making to fostering personalized treatment approaches. By analyzing this data, healthcare professionals can discern trends, identify risk factors, and optimize treatment protocols. Additionally, real-world data aids in post-market surveillance of drugs and medical devices, facilitating the detection of adverse events and enabling timely interventions. As the healthcare landscape evolves, harnessing the potential of real-world data promises to substantially elevate patient outcomes and refine the delivery of care.

In this work, the disease progression of Idiopathic pulmonary fibrosis (PF), a common interstitial lung disease with no cure\cite{ryu1998idiopathic} and Alzheimer’s Disease,  a progressive neurodegenerative disorder, were modeled. Pulmonary Fibrosis is a disease caused by the scarring of the lungs, making it more difficult for the lungs to work properly\cite{king2011idiopathic}. 
The disease does not progress at the same rate for all patients. 
For some patients, it progresses slowly, and live with PF for many years, while for others it declines more quickly. 
A key measure in assessing lung function to this end is forced vital capacity (FVC). 
This represents the total forced exhale volume (FEV) by a patient and is measured using a spirometer. 
A lower FVC value thus reflects a worse disease state for a fibrosis patient. 
In the evaluation of Alzheimer's Disease, the ADAS-13 (Alzheimer's Disease Assessment Scale-cognitive subscale with 13 items) serves as a pivotal metric for modeling cognitive impairment. An increase in the ADAS-13 scores is indicative of a deterioration in cognitive abilities, with higher scores corresponding to a more severe decline. This scale is instrumental in both clinical settings and research studies for monitoring the progression of the disease and assessing the impact of therapeutic interventions on cognitive function.

The current study aimed to apply Neural CDEs to model the disease progression of pulmonary fibrosis and Alzheimer’s Disease using multimodal (structural and image data) irregularly sampled data.
\section{Related Work}
 Chen et al. 2018 \cite{chen2018neural} reported a novel approach to deep learning by using ordinary differential equations (ODEs) as continuous-depth models. In this article, the authors demonstrated ODE-based models can outperform traditional ResNet \citenum{deng2009imagenet} architectures, offering improved efficiency and adaptability. Rubanova et al \cite{rubanova2019latent} have demonstrated ODE-basedmodels outperformed their RNN-based counterparts on irregularly-sampled data. Further to this study, Kidger et al, 2020 \cite{kidger2020neural} have developed the Neural CDE as a continuous analog of an RNN in sepsis prediction. In this study the authors introduced a new model for handling irregularly sampled time series data using neural controlled differential equations (Neural CDEs). The paper demonstrates that Neural CDEs can effectively model complex dynamics in irregular time series, outperforming traditional methods. The authors also highlight the potential of Neural CDEs in various applications, including healthcare and finance, where irregular time series are common. Wong et al, 2021 \cite{wong2021fibrosis} reported a specialized deep-learning model, Fibrosis-Net, designed to predict the progression of pulmonary fibrosis from chest CT images. This current work attempts to leverage the state-of-the-art Neural CDE approach in modeling time series data containing multiple modalities (structured data + image data).
\section{Methodology}

The Open-Source Imaging Consortium (OSIC) dataset\cite{ahmed2020osic} on interstitial lung diseases is a longitudinal dataset containing a baseline image at the 0th week, demographic information (age, gender, smoking status), and a series of measurements representing forced vital capacity (FVC). Key features of the data are that it includes 176 patients with a baseline CT scan and FVC data over 1.5 years. The dataset is longitudinal, containing a CT scan at week 0, which is the start of the study. From there, a series of forced vital capacity measurements (FVC) are recorded over the course of many weeks. FVC represents the total amount of air exhaled during a forced expiratory volume test as measured by a spirometer. FVC value can worsen as the patient experiences more and more shortness of breath over time, either through obstruction or restriction\cite{heckman2015pulmonary}. A decrease in FVC reflects disease progression.

The Alzheimer’s Disease dataset used in the preparation of this article were obtained from the Alzheimer’s Disease Neuroimaging Initiative (ADNI) database (adni.loni.usc.edu). ADNI dataset encompasses a wide array of information, including baseline and follow-up MRI and PET scans, demographic details (age, gender, education), genetic markers, and a variety of biomarkers from blood, urine, and cerebrospinal fluid samples. Key features of the data are that it comprises 1,566 patients across a spectrum of cognitive function, from normal aging to severe Alzheimer's disease. The dataset is longitudinal, with an initial scan and assessment serving as the baseline, followed by subsequent scans and evaluations at regular intervals. This allows for the tracking of neurobiological markers and cognitive metrics over time.

\subsection{Baseline Models}
The data was preprocessed to extract the relevant data and perform data wrangling. 
Two approaches for analyzing time-series data were evaluated - Neural CDE and LSTM. 
LSTMs model sequential data discretely through the use of a hidden state updated at every time step. 
Neural CDEs allow for modeling sequences in a continuous manner using differential equations allowing them to work well with irregularly sampled data. 
For analyzing image data (CT scans), the EfficientNet-b0 model was used\cite{tan2019efficientnet}. Other architectures were also evaluated, but in line with (Hamdi et al., 2019)\cite{hamdi2021marl}, this architecture served to be best in analyzing CT image slices.
As a next step, multimodal models were created by concatenating fixed-length output vectors produced from the image model with the latent embedding provided by the solved differential equation at each time step (or the cell output of an LSTM model), from which progression predictions were obtained with a last fully-connected layer. In both scenarios (baseline and attention-based models), a prediction approach of simulating disease progression for new, unobserved patients by only using their baseline CT image, time invariant data, and baseline FVC measurement were emplyed. This approach is analogous to a sequence-to-sequence task (using patients in the train dataset to predict disease trajectories for test patients) as opposed to a sequence-to-one approach where a patient's past history is used to forecast the next observation.

With regards to handling missing data with the Neural CDEs, the cubic spline/rectilinear interpolation (in addition to allowing for a continuous representation of the data) was used to fill in these values. For the LSTM, as is typical in the literature, a forward-fill approach was utilized.
\subsubsection{Structured Data}
Categorical features were label encoded and the continuous features and FVC label were normalized on the training dataset’s mean and standard deviation. For each patient progression, the data was shifted to use the next measured FVC value as the target. In addition, the week/month of the next measured FVC value was provided as an additional feature in the model to help inform the model's prediction.
\subsubsection{Multimodal Data} 
To test the hypothesis for the improvement in both LSTM and Neural-CDE performance, a baseline CT scan image was used. Similar preprocessing steps were performed on the structured data but for the multimodal data, each training step involved the incorporation of a random slice of the subject’s baseline CT scan. DICOM pixel intensities were rescaled, and the image was resized to a smaller size of 224 x 224 to fit on the GPU. A pretrained Efficient-Net model was used for image processing, and extracted features were then concatenated with LSTM/Neural CDE's features and passed through a final fully connected linear layer to give the final output.

\subsection{Baseline Model Architectures}
\subsubsection{Long-short term memory networks (LSTMs)}
Long-short term memory networks (LSTM), a specific type of Recurrent Neural Network (RNN), maintain state to model sequential data, using information from previous time steps to predict values at future time steps. LSTM uses various “gates” that add and remove information from the cell’s state. 

The equations that represent the computation performed in a single LSTM cell (a single LSTM layer at a particular time step) can be seen below:
$$ f_t = \sigma_g(\mathbf{W}_f x_t + \mathbf{U}_f h_{t-1} + b_f) $$
$$ i_t = \sigma_g(\mathbf{W}_i x_t + \mathbf{U}_i h_{t-1} + b_i) $$
$$ o_t = \sigma_g(\mathbf{W}_o x_t + \mathbf{U}_o h_{t-1} + b_o) $$
$$ c_t = f_t \cdot c_{t-1} + i_t \cdot \sigma_c(\mathbf{W}_c x_t + \mathbf{U}_c h_{t-1} + b_c) $$
$$ h_t = o_t \cdot \sigma_h(c_t) $$

For each cell, the cell state, and the hidden state from the previous step ($C_{t-1}$ and $h_{t-1}$) are utilized alongside input for the current time step ($x_t$) to create a new output ($o_t$) and a new cell and hidden state ($C_{t-1}$ and $h_{t-1}$)  via weights ($\mathbf{W}_f$, $\mathbf{W}_i$, $\mathbf{W}_o$, $\mathbf{W}_c$ and $\mathbf{U}_f$, $\mathbf{U}_i$, $\mathbf{U}_o$, $\mathbf{U}_c$) and biases ($b_f$, $b_i$, $b_o$, $b_c$) trained through backpropagation. LSTMs can furthermore be stacked such that inputs to the stacked LSTM are the hidden state at each time step from the previous LSTM network.

\begin{figure}[H]
    \centering
    \includegraphics[width=1\textwidth]{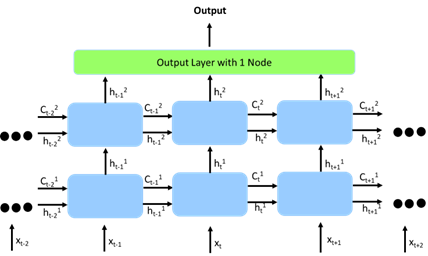}
    \caption{LSTM architecture}
    \label{fig:lstm}
\end{figure}

The baseline model used to compare against the Neural-CDE model was a stacked LSTM with two layers with a final fully connected layer with one layer (shown in Figure~\ref{fig:lstm}) to produce a single output: a prediction for FVC at the next observed time point.
\subsubsection{Neural Controlled Differential Equations (Neural CDE)}
\begin{figure}[H]
    \centering
    \includegraphics[width=0.3\textwidth]{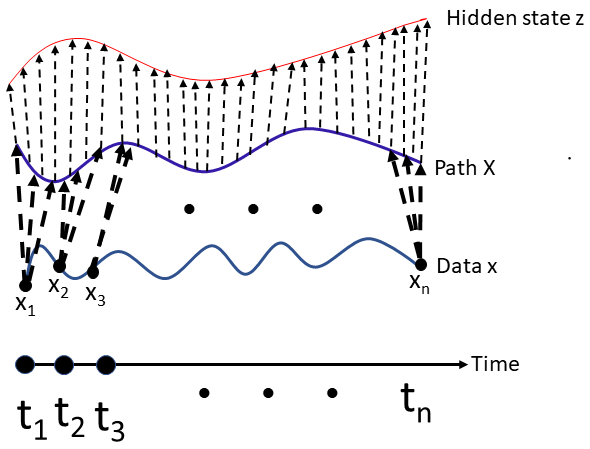}
    \caption{Neural CDE architecture}
    \label{fig:cde}
\end{figure}

Neural CDEs\cite{kidger2020neural} are a variant of NDEs designed to work specifically with irregularly sampled time series data. As shown in Figure~\ref{fig:cde}, Neural CDEs use a "time-varying vector" $\mathbf{X}$ to create a representation of the local dynamics from a calculated interpolation of the irregularly sampled data upon which a hidden state is calculated. A linear map is then used to make a final prediction. Mathematically, this can be represented by the following set of equations:
$$z(0)  = \zeta_\theta(t_0, x_0), z(t) = z(t_0) + \int_{t_0}^t f_{\theta_1}(z(s))\,\frac{\mathrm{d}{X}_\textbf{X}}{\mathrm{d}s}\mathrm{d}s$$ 

\subsection{Attention Based Model Architectures}

\subsubsection{Attetnion architecture}
\begin{figure}[hb!]
    \centering
    \includegraphics[width=0.6\textwidth]{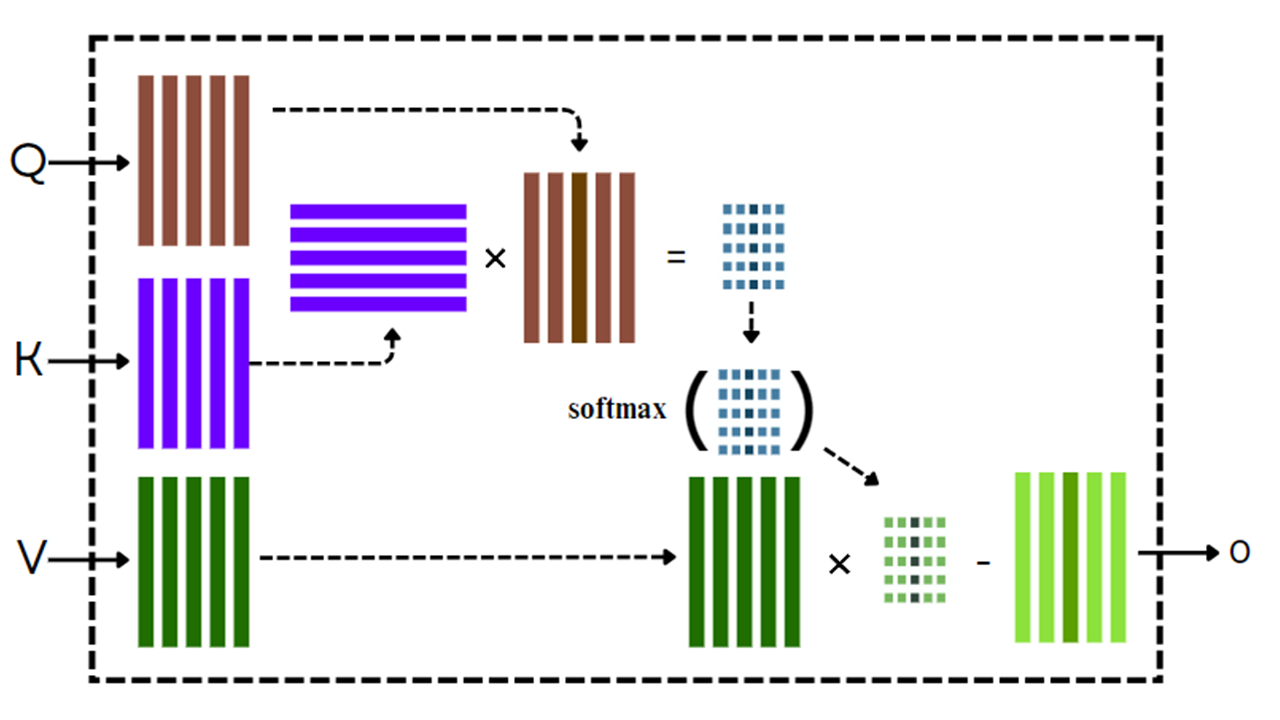}
    \caption{Illustration of attention operator}
    \label{fig:attention}
\end{figure}

The attention mechanism works by transforming the input data into Query (Q), Key (K), and Value (V) matrices10. The Q matrix represents the information that the model wants to retrieve from the input, while the K matrix represents the input itself. The model computes attention weights between Q and K, which are used to weight the values in the V matrix that are used to generate the output. The attention operator can be illustrated as shown in Figure~\ref{fig:attention}.

\subsubsection{Neural CDE Based Multimodal Multi-Head Fusion}

\begin{figure}[h]
    \centering
    \includegraphics[width=1\textwidth]{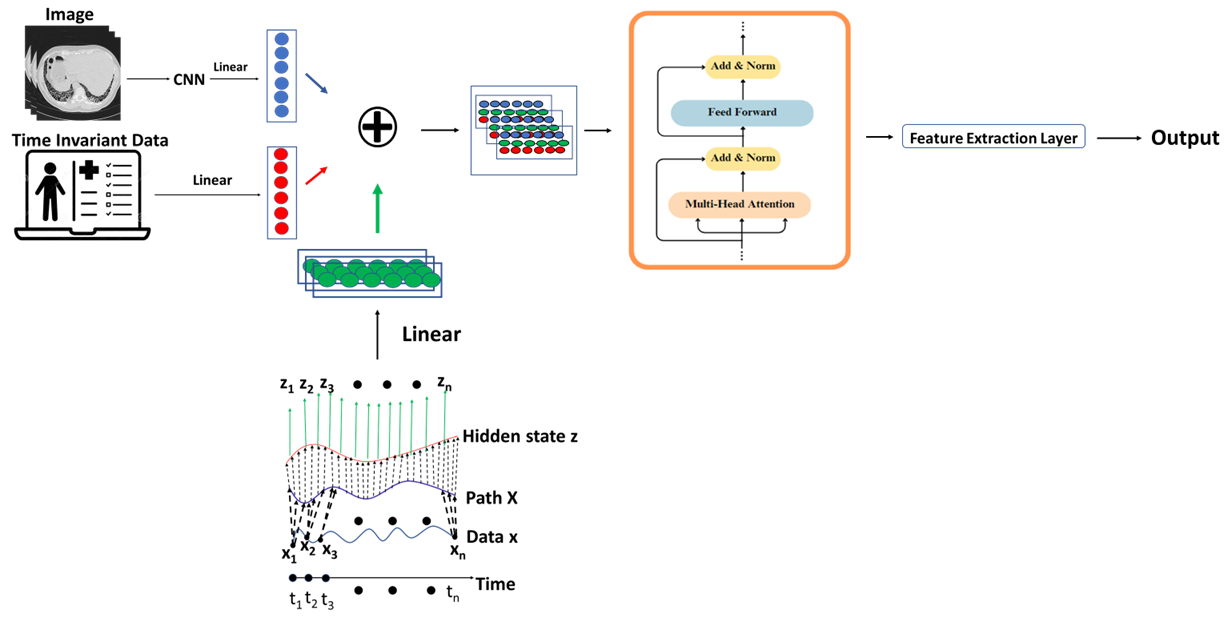}
    \caption{Neural CDE MultiModal Multi-Head Fusion Architecture}
    \label{fig:architecture}
\end{figure}

In order to incorporate multimodal context to help improve Neural CDE model performance through utilizing demographics information and image data, we introduce a novel methodology for incorporating Neural Differential Equation based models with multimodal deep learning-based models.

To incorporate multimodal context to aid the performance of Neural CDE model through the utilization of demographics information and image data, a novel methodology is introduced for the incorporation of Neural Differential Equation based models with multimodal deep learning-based models.

Initially, we utilize a rectilinear/Hermite spline interpolation with the Neural CDE, specifically chosen to ensure causality, ensuring that the interpolation at any given point is influenced solely by preceding time points, excluding future data.

Subsequently, the Neural CDE model equipped with a forecasting head —without any image data or demographic feature incorporation—is pretrained to learn initial trends in disease progression. This approach ensures that the subsequent multihead attention model will choose to more heavily incorporate the Neural CDE output early on in training as the Neural CDE model is more accurate in predicting disease trajectory. Over time, additional information from the multimodal context would be integrated. The model is evaluated at each time step a prediction is required at, and the model's coefficients are utilized by the forecasting head to make predictions. Following this pretraining step, the forecasting head is detached, and replaced with a linear map to a fixed length embedding vector representing the model's calculated features from the continuous hidden state. Embeddings are created from both the image slice data and the demographic details are then concatenated with CDE model's embeddings. Inspired by the Transformer's decoder and the masked multi-head attention we utilize a layer normalization  on the combined data and incorporate skip connections with a multihead attention mechanism that is constrained by a causal mask. Afterwards, we again apply layer normalization and subsequently incorporate skip connections in integrating a feedforward network to supplement the learned embedding so far. The resulting embeddings undergo processing via an intermediary hidden layer and culminate in a final linear mapping to provide the model's prediction for every individual timestep.

\begin{figure}[h]
    \centering
    \includegraphics[width=0.9\textwidth]{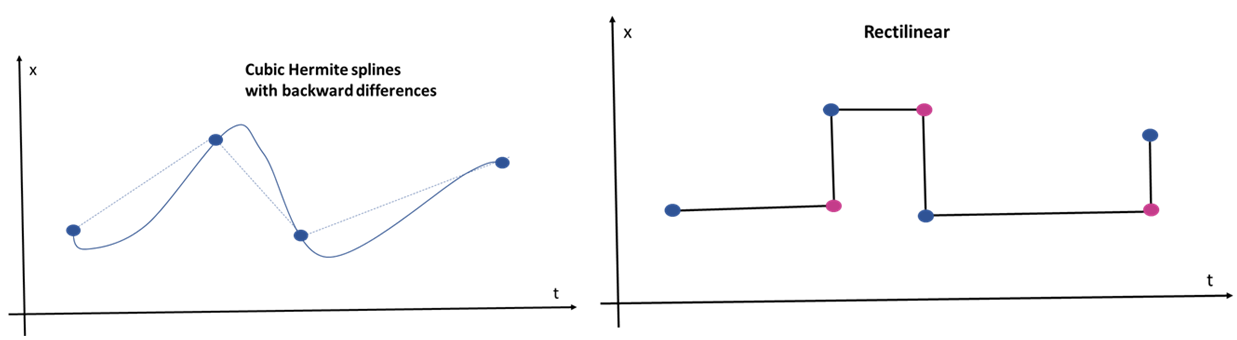}
    \caption{Neural CDE Data Interpolations}
    \label{fig:interpolation}
\end{figure}

\paragraph{Interpolation Schemes}
As shown in figure~\ref{fig:interpolation} we evaluated two different interpolation schemes (Cubic Hermite splines with backward differences and rectilinear interpolations) proposed in Morrill et al. to deal with the causality present in disease progression data. The main difference between the two approaches are that Cubic Hermite Splines are discretely online (can only be evaluated at time point at which new data is provided) whereas the Rectilinear interpolation is continuously online (model can be evaluated at any time). the Cubic Hermite Spline approach allows for discontinuities to be smoothened to allow for easier differential equation modeling. Additionally, the rectilinear approach, for which differential equation evaluation can be performed even beyond the irregularly sampled data can be seen as a generalization of the ODE-RNN time series modeling approach (Rubanova et al., 2019) \cite{rubanova2019latent}.

\subsubsection{LSTM Based Multimodal Multi-Head Fusion}
To further demonstrate the efficacy of the Neural CDE approach, we replace the Neural CDE model and its' interpolations with an additional time delta feature and forward filling any missing values. We then make use of the output of the model at each time step for subsequent embeddings with the image and demographic features.

\subsection{Model Training and Testing}
For all the models, the data was split in the following format: 70\% for training, 20\% for validation, 10\% as a holdout set. This was done to validate model performance. Root Mean Squared Error (RMSE), Mean Absolute Error (MAE), and R-Squared were calculated for training, validation, and test sets to assess the robustness of the model. Weights and Biases (wandb) was used for tracking and hyperparameter tuning to optimize the model performance.
\section{Results and Discussion}
\subsection{Baseline Model Architectures}
\subsubsection{Structural data only versus Multimodal Data}

As shown in Table~\ref{tab:struct_vs_multi}, using the structured data alone, the Neural CDE model achieves relatively good performance with a low training RMSE of 1.076, a validation RMSE of 1.054, and a test RMSE of 1.215. These results suggest that the model can capture and generalize the underlying patterns within the structured data. In the case of multimodal data, the Neural CDE model performs better with the training RMSE lower at 0.4147, indicating the model's ability to effectively learn from the diverse data sources. The validation RMSE is 0.6559, further affirming the model's robustness. The lowest test RMSE of 0.5405 highlights the Neural CDE's capability to generalize well to unseen multimodal data.

\begin{table}[!htp]
\centering

\sisetup{detect-all}
\NewDocumentCommand{\B}{}{\fontseries{b}\selectfont}

\begin{tabular}{
  @{}
  l
  S[table-format=1.2]
  S[table-format=1.2]
  S[table-format=1.2]
  S[table-format=1.2]
  S[table-format=-1.2]
  S[table-format=1.2]
  S[table-format=1.2]
  S[table-format=1.2]
  S[table-format=1.2]
  @{}
}
\toprule
Neural CDE Models 
\\ 
& {Train RMSE} & {Validation RMSE} & {Test RMSE} \\ 
\midrule
Structured Data & 1.076 &  1.054 &  1.215 \\ 
Multimodal Data & 0.4147 & 0.6559 & \B 0.5405 \\ 
\bottomrule
\end{tabular}

\caption{Performance of Neural CDE model using structured data alone versus multimodal data}
\label{tab:struct_vs_multi}
\end{table}

\subsubsection{Multimodal LSTM versus Multimodal Neural CDE}
As shown in Table \ref{tab:multi_metrics}, the multimodal LSTM model exhibits good performance  across  the training phase with a train RMSE of 0.5824. However, its validation and test RMSE values are comparatively higher at 0.8586 and 1.396, respectively. This indicates that while multimodal LSTM captures certain patterns during training, it might struggle to generalize well to unseen data, resulting in higher error rates. On the other hand, the Neural CDE model performs better as it achieves a significantly lower train RMSE of 0.4147 and validation RMSE of 0.6559. The test RMSE of Neural CDE 0.5405 is the lowest among the evaluated models. This suggests that Neural CDE better in capturing the training patterns but also in generalizing effectively to new instances.

\begin{table}[!htp]
\centering

\sisetup{detect-all}
\NewDocumentCommand{\B}{}{\fontseries{b}\selectfont}

\begin{tabular}{
  @{}
  l
  S[table-format=1.2]
  S[table-format=1.2]
  S[table-format=1.2]
  S[table-format=1.2]
  S[table-format=-1.2]
  S[table-format=1.2]
  S[table-format=1.2]
  S[table-format=1.2]
  S[table-format=1.2]
  @{}
}
\toprule
Models 
\\ 
& {Train RMSE} & {Validation RMSE} & {Test RMSE} \\ 
\midrule
LSTM & 0.5824 &  0.8586 &  1.396 \\ 
Neural CDE & 0.4147 & 0.6559 & \B 0.5405 \\ 
\bottomrule
\end{tabular}

\caption{Model performance of multimodal LSTM versus multimodal Neural CDE}
\label{tab:multi_metrics}
\end{table}

\subsection{Attention Based Architectures}
\subsubsection{Fusion strategies}
As shown in Table \ref{tab:fusion_metrics}, the Embedding Sum method demonstrates a reasonable performance across different phases. In the training phase, it achieves an RMSE of 0.8879, and its performance slightly deteriorates in the validation phase, with an RMSE of 1.118. However, it manages to maintain its effectiveness in the testing phase with an RMSE of 0.902. In contrast, the Embedding Concatenation technique stands out with a notably lower RMSE across all phases. It performs well with an RMSE of 0.1360 in the training phase, maintains its performance in the validation phase with an RMSE of 0.3465, and achieves the best result in the testing phase with an RMSE of 0.2912. We observed information loss when we used addition as the fusion strategy - loss stabilized at a much higher error than we observed with the concatenation approach. These findings suggest the benefit of using the Embedding Concatenation approach in minimizing prediction errors compared to the Embedding Sum method. 

\begin{table}[!htp]
\centering

\sisetup{detect-all}
\NewDocumentCommand{\B}{}{\fontseries{b}\selectfont}

\begin{tabular}{
  @{}
  l
  S[table-format=1.2]
  S[table-format=1.2]
  S[table-format=1.2]
  S[table-format=1.2]
  S[table-format=-1.2]
  S[table-format=1.2]
  S[table-format=1.2]
  S[table-format=1.2]
  S[table-format=1.2]
  @{}
}
\toprule
Models 
\\ 
& {Train RMSE} & {Validation RMSE} & {Test RMSE} \\ 
\midrule
Sum & 0.8879 &  1.118 &  0.902 \\ 
Concatenation & 0.1360 & 0.3465 & \B 0.2912 \\ 
\bottomrule
\end{tabular}

\caption{Comparison of the neural CDE model performance with embedding using sum versus concatenation}
\label{tab:fusion_metrics}
\end{table}

\subsubsection{LSTM fusion versus  Neural CDE fusion}

As shown in Table \ref{tab:realdiffusionnet}, the LSTM Fusion model demonstrates a consistent performance pattern across different phases of the process. During training, it achieves an RMSE of 0.3038, maintaining its effectiveness in the validation phase with an RMSE of 0.8191. The performance remains stable in the testing phase as well, with an RMSE of 0.9066. In comparison, the Neural CDE Fusion approach outperforms the LSTM Fusion model in terms of predictive accuracy. It starts with a lower RMSE of 0.1360 during training, which carries over into the validation phase with an RMSE of 0.3465. Notably, the Neural CDE Fusion model achieves the best result in the testing phase, demonstrating an impressive RMSE of 0.2912. These findings highlight the superiority of the Neural CDE Fusion model in minimizing prediction errors compared to the LSTM Fusion approach.

\begin{table}[!htp]
\centering

\sisetup{detect-all}
\NewDocumentCommand{\B}{}{\fontseries{b}\selectfont}

\begin{tabular}{
  @{}
  l
  S[table-format=1.2]
  S[table-format=1.2]
  S[table-format=1.2]
  S[table-format=1.2]
  S[table-format=-1.2]
  S[table-format=1.2]
  S[table-format=1.2]
  S[table-format=1.2]
  S[table-format=1.2]
  @{}
}
\toprule
Models 
\\ 
& {Train RMSE} & {Validation RMSE} & {Test RMSE} \\ 
\midrule
LSTM Fusion & 0.3038 &  0.8191 &  0.9066 \\ 
Neural CDE Fusion & 0.1360 & 0.3465 & \B 0.2912 \\ 
\bottomrule
\end{tabular}

\caption{Comparison of performance LSTM fusion versus Neural CDE fusion}
\label{tab:realdiffusionnet}
\end{table}

\subsubsection{Neural CDE interpolation strategies}

As shown in Table \ref{tab:spline}, the Cubic Hermite splines model exhibits consistent results throughout the process. During training, it achieves an RMSE of 0.1360, maintaining its effectiveness in the validation phase with an RMSE of 0.3465. The model's performance remains stable in the testing phase as well, with an RMSE of 0.2912. In contrast, the Rectilinear interpolation approach outperforms the Cubic Hermite splines model in terms of predictive accuracy. It starts with a slightly lower RMSE of 0.1278 during training, which carries over into the validation phase with an RMSE of 0.3371. Notably, the Rectilinear interpolation model achieves the best result in the testing phase with an RMSE of 0.2570. These findings highlight the advantage of the Rectilinear interpolation model in minimizing prediction errors compared to the Cubic Hermite splines approach. 

\begin{table}[!htp]
\centering

\sisetup{detect-all}
\NewDocumentCommand{\B}{}{\fontseries{b}\selectfont}

\begin{tabular}{
  @{}
  l
  S[table-format=1.2]
  S[table-format=1.2]
  S[table-format=1.2]
  S[table-format=1.2]
  S[table-format=-1.2]
  S[table-format=1.2]
  S[table-format=1.2]
  S[table-format=1.2]
  S[table-format=1.2]
  @{}
}
\toprule
Models 
\\ 
& {Train} & {Validation} & {Test} \\ 
\midrule
Cubic Hermite Splines & 0.1360 & 0.3465  &  0.2912 \\ 
Rectilinear & 0.1278 & 0.3371 & \B 0.2570 \\ 
\bottomrule
\end{tabular}

\caption{Model performance with the use of Cubic Hermite splines versus Rectilinear Interpolation}
\label{tab:spline}
\end{table}

\subsubsection{Pretrained Neural CDE versus RealDiffFusionNet}

As shown in Table \ref{tab:PreVsFusion}, the pretrained CDE model displays consistent results throughout the process. During training, it achieves an RMSE of 1.076, maintaining its effectiveness in the validation phase with an RMSE of 1.054. The model's performance remains relatively stable in the testing phase as well, with an RMSE of 1.215. In contrast, the RealDiffFusionNet approach outperforms the pretrained CDE model in terms of predictive accuracy. It starts with a significantly lower RMSE of 0.1278 during training, which carries over into the validation phase with an RMSE of 0.3371. Notably, the RealDiffFusionNet achieves the best result in the testing phase, demonstrating a lower RMSE of 0.2570. These findings highlight the superiority of the RealDiffFusionNet in minimizing prediction errors compared to the pretrained CDE approach.

\begin{table}[!htp]
\centering

\sisetup{detect-all}
\NewDocumentCommand{\B}{}{\fontseries{b}\selectfont}

\begin{tabular}{
  @{}
  l
  S[table-format=1.2]
  S[table-format=1.2]
  S[table-format=1.2]
  S[table-format=1.2]
  S[table-format=-1.2]
  S[table-format=1.2]
  S[table-format=1.2]
  S[table-format=1.2]
  S[table-format=1.2]
  @{}
}
\toprule
Models 
\\ 
& {Train RMSE} & {Validation RMSE} & {Test RMSE} \\ 
\midrule
Pretrained CDE & 1.076 &  1.054 &  1.215 \\ 
RealDiffFusionNet & 0.1278 & 0.3371 & \B 0.2570 \\ 
\bottomrule
\end{tabular}

\caption{Comparison of model performance with pretrained Neural CDE versus RealDiffFusionNet}
\label{tab:PreVsFusion}
\end{table}

\section{Conclusions}
This study evaluated the performance of Neural CDE informed multihead attention models in comparison to various baseline models, including LSTMs. It was found that the use of the Neural CDE based models served to improve accuracy due to their capacity to handle the irregularly sampled medical data. Additionally, it was observed that the incorporation of the multimodal context (baseline CT scan and patient time invariant features) using attention fusion was found to be successful in improving model performance. Overall, ReallDiffFusionNet is promising in analyzing real world healthcare time series data.
 \newpage


\bibliographystyle{IEEEtran}
\bibliography{sample-base}

\end{document}